\documentclass[utf8]{frontiersSCNS} % for Science, Engineering and Humanities and Social Sciences articles
\usepackage[T1]{fontenc}
\usepackage{url,hyperref,lineno,microtype,subcaption}
\usepackage[onehalfspacing]{setspace}
\usepackage{multirow}
\usepackage{booktabs}
\usepackage{array}
\usepackage{graphicx}
\usepackage{epstopdf}

\def\keyFont{\fontsize{8}{11}\helveticabold }
\def\firstAuthorLast{Xiang {et~al.}} %use et al only if is more than 1 author
\def\Authors{Yun Xiang\,$^{1}$, Qijun Chen\,$^{1}$, Zhongjin Su\,$^{1}$, Lu Zhang\,$^{1}$, Zuohui Chen\,$^{1}$, Guozhi Zhou\,$^{2}$, Zhuping Yao\,$^{2}$,Qi Xuan\,$^{1,*}$ and Yuan Cheng\,$^{2,*}$}
\def\Address{$^{1}$Institute of Cyberspace Security, Zhejiang University of Technology, Hangzhou, China \\
$^{2}$Institute of Vegetables, Zhejiang Academy of Agricultural Sciences, Hangzhou, China  }
\def\corrAuthor{Qi Xuan, Yuan Cheng}

\def\corrEmail{xuanqi@zjut.edu.cn, Chengyuan1005@126.com}

\begin{document}
\onecolumn
\firstpage{1}

\title[Hyperspectral Imaging for cherry tomato]{Deep Learning and Hyperspectral  Images Based Tomato Soluble Solids Content and Firmness Estimation}

\author[\firstAuthorLast ]{\Authors} %This field will be automatically populated
\address{} %This field will be automatically populated
\correspondance{} %This field will be automatically populated

\extraAuth{}% If there are more than 1 corresponding author, comment this line and uncomment the next one.
%\extraAuth{corresponding Author2 \\ Laboratory X2, Institute X2, Department X2, Organization X2, Street X2, City X2 , State XX2 (only USA, Canada and Australia), Zip Code2, X2 Country X2, email2@uni2.edu}

\maketitle

\begin{abstract}

%%% Leave the Abstract empty if your article does not require one, please see the Summary Table for full details.
\section{}
Cherry tomato (Solanum lycopersicum) is popular with consumers over the world due to its special flavor. Soluble solids content (SSC) and firmness are two key metrics for evaluating the product qualities. In this work, we develop non-destructive testing techniques for SSC and fruit firmness based on hyperspectral images and corresponding deep learning regression model. Hyperspectral reflectance images of over 200 tomato fruits are derived with spectrum ranging from 400 to 1000 nm. The acquired hyperspectral images are corrected and the spectral information are extracted. A novel one-dimensional(1D) convolutional ResNet (Con1dResNet) based regression model is prosed and compared with the state of art techniques. Experimental results show that, with relatively large number of samples our technique is 26.4\% better than state of art technique for SSC and 33.7\% for firmness. The results of this study indicate the application potential of hyperspectral imaging technique in the SSC and firmness detection, which provides a new option for non-destructive testing of cherry tomato fruit quality in the future.

\tiny
 \keyFont{ \section{Keywords:} hyperspectral imaging, deep learning, cherry tomato, soluble solids content, One-dimensional Convolutional Neural Networks, firmness} %All article types: you may provide up to 8 keywords; at least 5 are mandatory.
\end{abstract}

\section{Introduction}

Tomato is a very popular fruit globally and its annual production reaches 186.82 million tons in 2020\citep{Fao}. Tomatoes contain rich nutrients such as lycopene, $\beta$-carotene and vitamins ~\citep{sainju2003mineral,gao2020wrky} etc. To facilitate the tomato production, processing, and marketing, its grade and maturity needs to be evaluated. In general, soluble solids and firmness are two key indicators~\citep{beckles2012factors}. SSC can be used to grade tomato quality and the firmness can be used to determine fruit maturity ~\citep{peng2008analysis}. The existing measuring techniques relying upon chemistry reactions can derive the SSC value accurately. However, the destructive methods can not be applied in high volume measurements.  Moreover, there are significant variations so that sampling can be inefficient and ~inaccurate\citep{li2013comparative}. Therefore, in this work, we propose a hyperspectral imaging and deep learning based technique to measure tomato SSC and firmness nondestructively, accurately, and in high volume.

Spectroscopy is a widely used nondestructive testing method for fruit inspection. It includes various imaging techniques including visible, near infrared, terahertz spectroscopy, raman spectroscopy, and hyperspectral imaging etc. Visible and near infrared spectroscopy are rapid, convenient, and low cost. However, they are contrained by limited spectral band\citep{yin2019review}. Terahertz (THz) radiation has microwave and infrared properties and is able to penetrate and interact with many common materials, its equipments are very expensive\citep{afsah2019comprehensive}. Raman spectroscopy is easy to operate, quick to measure, and contains rich information. However, its performance is inferior in terms of stability and sensitivity~\citep{weng2019recent}. Hyperspectral imaging technology can simultaneously detect the two-dimensional spatial information and 1D spectral information, therefore combine image and spectral characteristics~\citep{adao2017hyperspectral}. It can derive the overall spatial spectral information of cherry tomato and thus, is selected as the imaging method.

Hyperspectral imaging has been widely used for non-destructive testing in various fields, such as detection of plant disease stress \citep{lowe2017hyperspectral}, industrial food packaging \citep{medus2021hyperspectral}, medical image classification \citep{jeyaraj2019computer}, and horticultural products \citep{huang2017development}. Hyperspectral images are also effective for quality analysis of fruits.  \cite{rahman2017nondestructive} use hyperspectral imaging to estimate metrics such as water content and PH readings. \cite{zhou2020non} use it to classify the maize seeds. \cite{fan2015prediction} use it to predict SSC and firmness in pears. They combine the competitive adaptive reweighted sampling and successive projection algorithm to select the variables as in partial least squares regression(PLSR).  \cite{rahman2018hyperspectral} fit sweetness and firmness of  tomato. \cite{lu2017innovative} gives a review of the application of recent hyperspectral techniques. Therefore, hyperspectral imaging techniques can effectively measure or classify fruit and vegetable products.

The existing spectral analysis techniques typically require a regression model to fit the spectral data ~\citep{jiang2015chemometric}, which have been widely used in areas such as food, petrochemical, and pharmaceutical fields \citep{chen2018application}. In general, various machine learning based algorithms are employed to build classification and regression models for hyperspectral images. \cite{li2016application} use PLSR to build a hyperspectral regression model to predict the water status of grapevines. \cite{guo2016hyperspectral} develop an SVM model to assess the maturity of strawberries. \cite{abdulridha2019uav} combine hyperspectral imaging and KNN algorithm to differentiate ulcer-infected fruits. \cite{ji2019detection} use the AdaBoost algorithm to recognize the rate of potato damage. The machine learning algorithms typically perform a filtering process on the spectral bands.

Deep learning models, e.g., convolutional neural network (CNN), can learn features automatically from a large amount of data \citep{guo2016deep}. It is widely used in medics \citep{esteva2019guide}, industry \citep{hossain2018automatic}, agriculture \citep{kamilaris2018deep}, object detection \citep{zou2019object}, and signal processing \citep{yu2010deep} etc. This technique is also used in building hyperspectral correction models for classification and prediction. \cite{paoletti2019deep} summarize the application of deep learning for hyperspectral image classification and  conclude that CNN based models are generally more effective due to their capacity to extract highly discriminatory features and leverage the spatial and spectral information. \cite{qiu2018variety} demonstrate that CNN outperforms other machine learning methods for rice variety identification application. \cite{kong2014fast} track activity of peroxidase in tomato hyperspectral images using genetic algorithm and extreme learning machine. \cite{rahman2018hyperspectral} develop a regression model in 1000- 1550 nm hyperspectral images using PLSR method to estimate sweetness and firmness with ${R^2}$ of 0.672 and 0.548, respectively.

In this work, we propose a deep learning and hyperspectral imaging based technique to estimate the metrics inside cherry tomato. Specifically, we have made the following contributions.
\begin{enumerate}
\item  We demonstrate the effectiveness of deep learning based techniques and propose such a model to estimate fruit SSC and firmness.
\item  We explore the tradeoff between sample number and model accuracy.
\item  We collect real-world field data and evaluate the performance of our technique.
\end{enumerate}

The experimental results show that our technique is 26.4\% better than the state of art technique in SSC estimation and 33.4\% in firmness estimation.

\section{MATERIALS AND METHODS}

In this section, we describe the sample preparation, hyperspectral image acquisition and calibration, and the ground truth measurements for SSC and firmness methods. Specifically, we develop Con1dResNet, a deep learning and  hyperspectral image based  SSC and firmness estimation technique. Meanwhile, four comparing baseline techniques are also introduced.

\subsection{Sample Preparation}

The sample plant is a local mainstream cherry tomato (cultivar: Zheyingfen-1). The seeds first grow in the lab with tight environment control for one month. Then the seedlings are transplanted to the greenhouse of the Zhejiang academy of agricultural sciences, Hangzhou, China (east longitude 120$^{\circ}$2', north latitude 30$^{\circ}$27') on April 2nd (early spring), 2021. Field management is implemented following the standard commercial procedures. Cherry tomato fruits are harvested in June 2021. 200 fully mature fruits are collected from 50 different plants for hyperspectral image acquisition. Firmness and soluble solids content of each fruit is measured using portable firmness tester and hand-held refractometer after image acquisition, respectively.

\subsubsection{Hyperspectral Image Acquisition}

\begin{figure}[h!]
\begin{center}
\includegraphics[width=10cm]{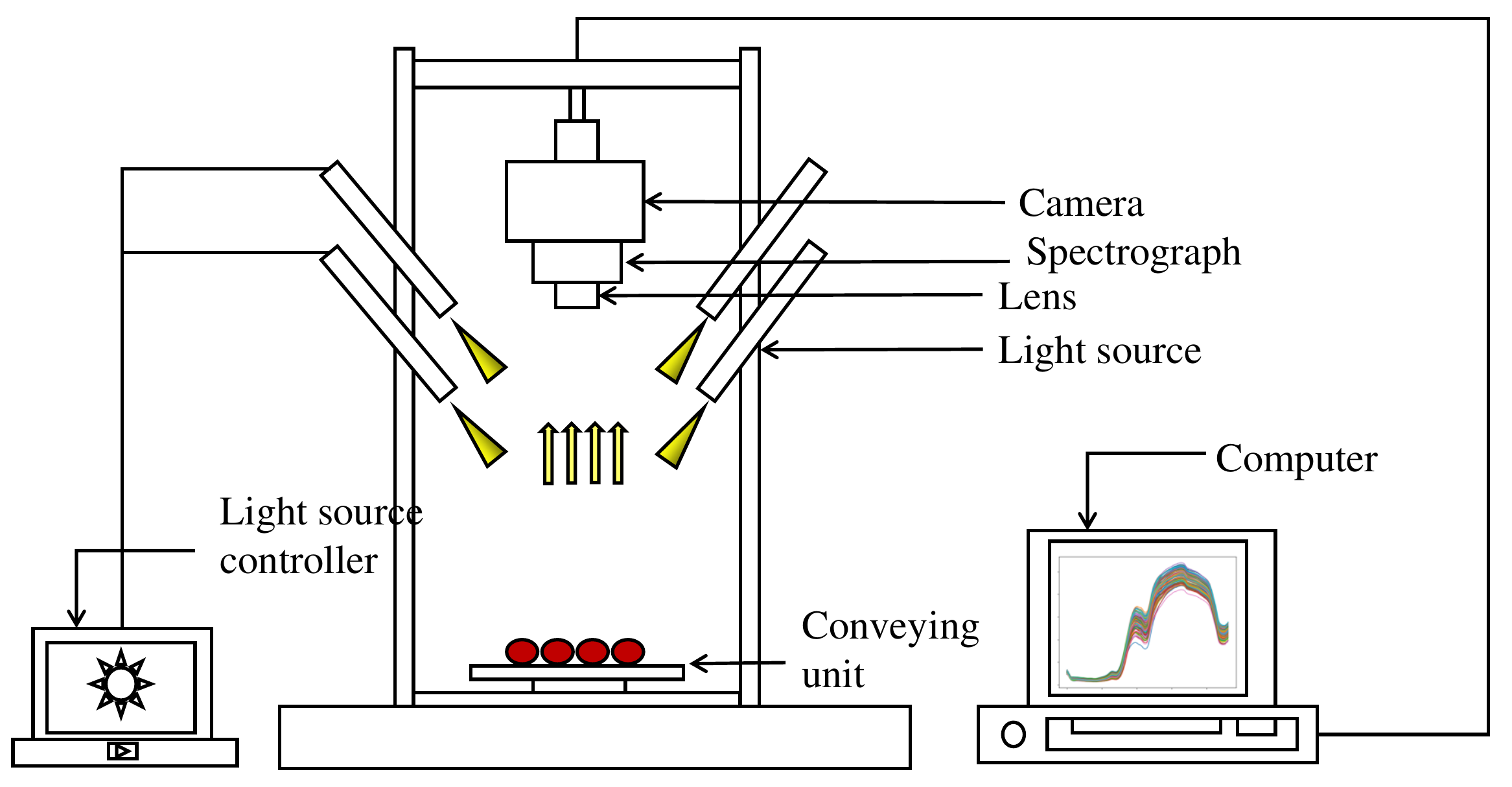}% This is a *.eps file
\end{center}
\caption{ Schematic of the hyperspectral imaging system for acquiring spectral scattering images from cherry tomatoes}\label{fig:1}
\end{figure}

\begin{figure}[h!]
	\begin{center}
		\includegraphics[width=10cm]{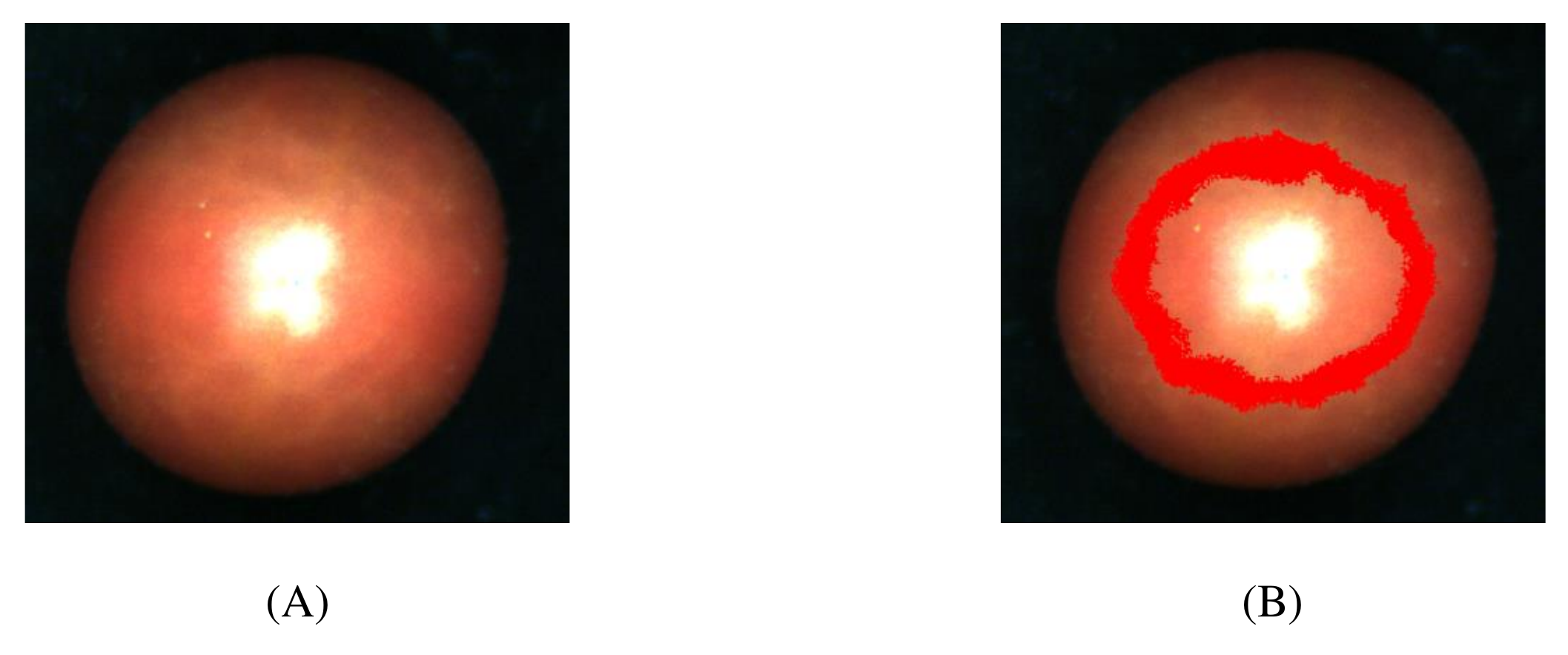}% This is a *.eps file
	\end{center}
	\caption{ \textbf{(A)} ENVI original hyperspectral image, \textbf{(B)} area map of ROI acquired by ENVI}\label{fig:2}
\end{figure}

A hyperspectral imaging system is used to derive the hyperspectral images as shown in Fig.\ref{fig:1}. We use a push-broom hyperspectral camera (PIKA XC, Resonon Inc., Bozeman, MT, USA) mounted 20 cm above the tomato samples. The hyperspectral images are acquired with the spatial resolution of 50 pixels per ${mm^2}$ under artificial lighting (four 15W 12 V light bulbs with two on either side of the lens). The main specifications of the hyperspectral camera were: interface, Firewire (IEEE 1394b), digital output (14 bit), and angular field of view of 7 degrees. The objective lens had a 17 mm focal length (maximum aperture of F1.4), optimized for the  hyperspectral. We acquire reflectance data in 462 spectral bands from 386 to 1004 nm with a spectral resolution of 1.3 nm. Due to the convex surface of the samples, the uneven reflection creates a highlighted region near the vertical axial as shown in Fig.\ref{fig:2}A. Thus, we use ENVI5.3 (ITT, Visual Information Solutions, Boulder, CO, USA) \citep{su2021application} to avoid the highlight region and extract the reflection value for each band from the region of interest \citep{xue2010application,fu2021nondestructive} (Fig.\ref{fig:2}B).

%A multispectral imaging system is used for this research, which is shown in Fig.\ref{fig:1}, and its detailed parameters are shown in Table \ref{tab:1}. We used a push-broom hyperspectral camera (PIKA XC, Resonon Inc., Bozeman, MT, USA) mounted 20 cm above the specimens. The hyperspectral images were acquired with the spatial resolution of 50 pixels per ${mm^2}$ under artificial lighting (four 15W 12 V light bulbs with two on either side of the lens). The main specifications of the hyperspectral camera were: interface, Firewire (IEEE 1394b), digital output (14 bit), and angular field of view of 7 degrees. The objective lens had a 17 mm focal length (maximum aperture of F1.4), optimized for the near-infrared and visible near-infrared waveband
%We acquired reflectance data in 462 spectral bands from 386 to 1004 nm (spectral resolution = 1.3 nm). Due to the convex surface of the cherry tomato samples, the reflectance of the sample surface is uneven, resulting in a highlight region in the image of each sample (Fig.\ref{fig:2}A). This study using ENVI5.3 (ITT, Visual Information Solutions, Boulder, CO, USA) excluded the highlight region to extract the region of interest \citep{fu2021nondestructive} (Fig.\ref{fig:2}B).
% For Original Research articles, please note that the Material and Methods section can be placed in any of the following ways: before Results, before Discussion or after Discussion.

\subsubsection{Hyperspectral Image Calibration}

\begin{figure}[h!]
	\begin{center}
		\includegraphics[width=15cm]{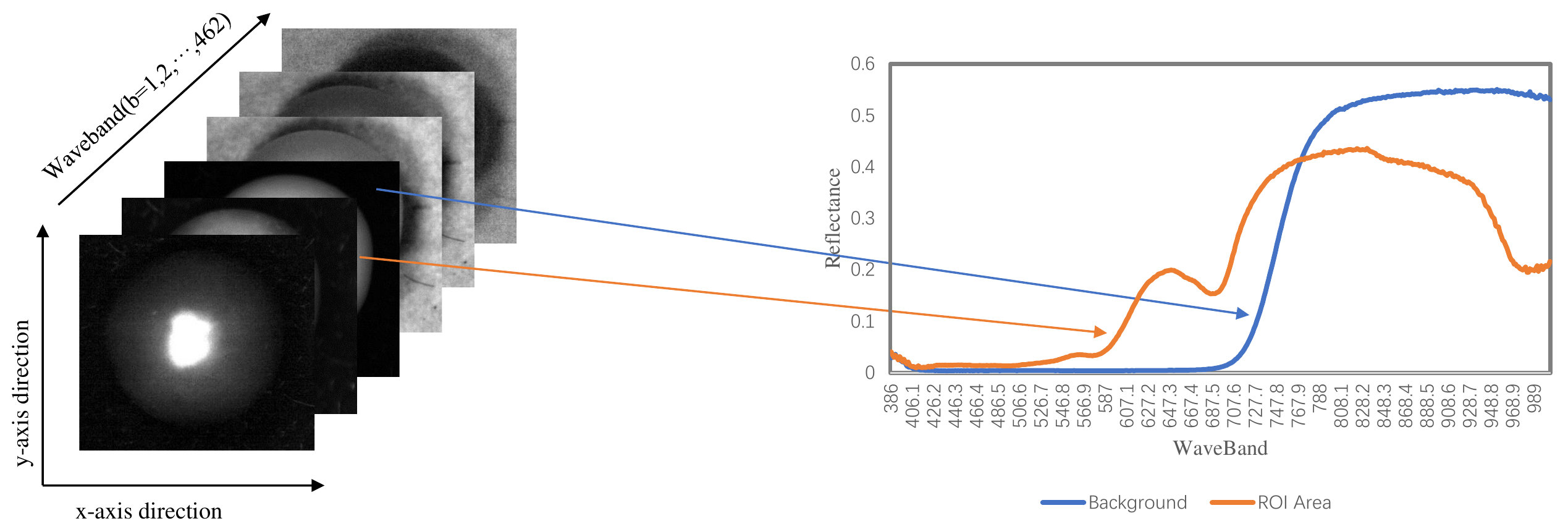}% This is a *.eps file
	\end{center}
	\caption{Schematic diagram of the structure and data of the corrected hyperspectral image: spatial axis x, y and wavebands}\label{fig:3}
\end{figure}

In reflectance calibration, the acquired hyperspectral image needs to be calibrated for the background spectral response of the instrument and the thermal dark current of the camera. The spectral data collected from the CCD device contains only the detector signal intensity value~\citep{elmasry2012principles}. Therefore, it is required to convert the raw data to reflectance or absorptivity values by comparing to the spectra of standard reference substances\citep{burger2005hyperspectral} as shown in Fig.\ref{fig:3}. The reflectance can be derived using the following equation.

$$
R_{c}=\frac{R_{ori}-R_{dark}}{R_{white}-R_{dark}},
$$
where ${R_c}$ is the corrected hyperspectral reflectance, ${R_{ori}}$ is the original reflection value extracted from ENVI5.3, ${R_{dark}}$ is the dark environment hyperspectral image reflection value, which is acquired using an opaque lens cap covering the hyperspectral lens, and ${R_{white}}$ is the reflection value of a piece of white Teflon (100\% reflectance, K-Mac Plastics, MI, USA).

\subsubsection{Baseline Measurement}

The baseline firmness and SSC of cherry tomatoes are measured in the lab. For the firmness measurement, the cherry tomatoes are fixed on a portable firmness measurement equipment (GY-4, Zhejiang Top Cloud-Agri Technology Co., Ltd, China). The equipment is zero-calibrated. Starting from the contact of the probe with the cherry tomato surface, the 10 mm downward pressure is considered as the firmness value.

SSC measurements follow the firmness measurements. Cherry tomatoes are cut along the vertical axis and wrapped using a gauze. Then they are squeezed manually to force out the solution. About one milliliter tomato solution is placed on the prism of a portable digital refractometer (PAL-1, ATAGO CHINA Guangzhou Co., Ltd, China) to derive the baseline SSC readings. Each cherry tomato sample solution is measured for 3 times and the results are averaged to reduce the effect of random environment events.

\subsection{Image Processing Models}

\subsubsection{Deep Learning Model}

Deep learning models are widely used in medical image processings \citep{kiranyaz2015convolutional}. However, in this work, it is required to build appropriate regression models. In general, we propose the Con1dResNet model to estimate the tomato SSC and firmness. 

ResNet ~\citep{he2016identity}, a popular model for image classification, can solve the degradation problem of deep networks. Thus, ResNet34 is implemented as the baseline network structure, and the original convolutional layer is reconstructed to be one-dimensional, accordingly. We use the Adam optimizer and  mean squared error loss function. We change the number of categories output by the last fully connected layer to one so that the network directly outputs the estimated values of SSC and firmness.

\begin{figure}[h!]
	\begin{center}
		\includegraphics[width=16cm]{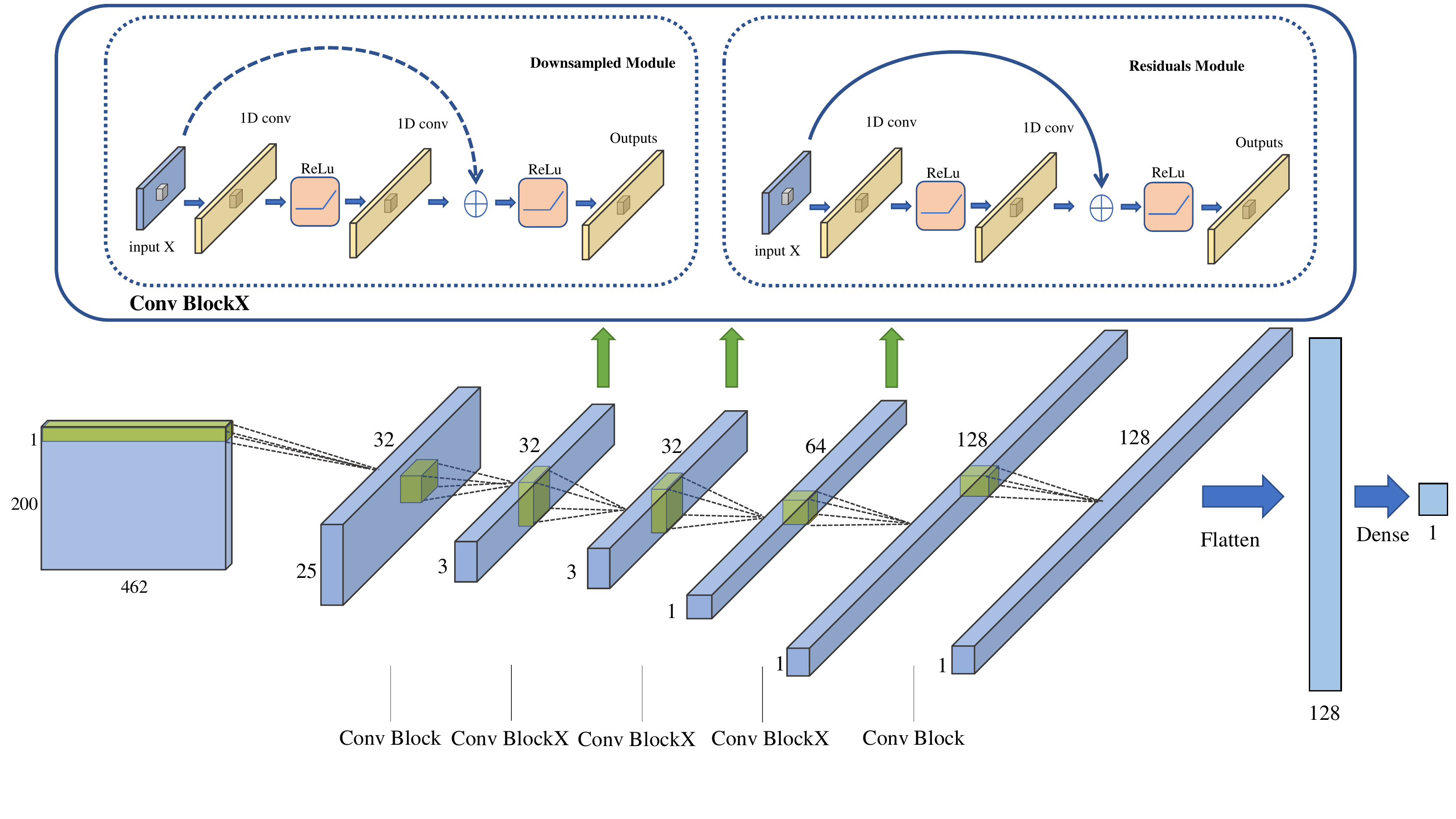}% This is a *.eps file
	\end{center}
	\caption{Con1dResNet network structure schematic}\label{fig:4}
\end{figure}

The specific network structure is shown in Fig.\ref{fig:4}. In the figure, the input is the reflectance values of the processed 462 spectral bands. There are five main blocks. The first block consists of a 1D convolution layer and a maximum pool layer, and then continues through a dropout layer with parameter 0.5. The second blockX contains 3 residuals module. The third blockX contains 1 downsampled module and 3 residuals module. The fourth blockX goes through 1 downsampled module and 5 residuals module before a dropout layer with parameter 0.5, and then continues through 3 residuals module. The fifth block consists of a mean pool layer and linear output layer. The number of convolution filters doubles as the block goes deeper (starting with 32 and ending with 128). All convolutional layers have a kernel size of 3 and a step size of 3. By connecting the convolutional layers together, deeper layers can be connected to a larger portion of the original input. Thus, different layers see the original input and learning ability at different levels. The last deeper layer outputs the SSC estimation, which converge to the ground truth value under the approximation of the MSE loss function.

\subsubsection{Machine Learning Models}

The existing image feature extraction methods are mostly based on machine learning models. In this work, we select four widely-used models as references to our deep learning based technique.

\paragraph{SVR}

Support Vactor Regression(SVR) \citep{castro2009online} is a variant of the support vector machines (SVM) technique~\citep{burges1998tutorial}. SVR is widely used in spectral data analysis for its linear and nonlinear processing ability. SVR maps the data to a high-dimensional space and constructs a hyperplane to minimize the distance to the farthest sample point. In SVR, the selection of an appropriate kernel function is crucial. We implement an SVR model with Radial basis function (RBF), which can process nonlinear data effectively.

\paragraph{KNNR}

K-Nearest Neighbors Regression(KNNR) \citep{yao2006regression}  is a nonparametric method that approximates the association between independent variables and continuous outcomes. Intuitively, it averages observations in the same neighborhood. The size of the neighborhood can be set using cross-validation to minimize the mean squared error. The number of neighborhoods in this work is set to 5.

\paragraph{AdaBoostR}

Adaptive Boosting Regression(AdaBoostR) \citep{freund1999short} is an iterative boosting algorithm. An AdaBoost regressor is a meta-estimator. It first fits a regressional measure on the original dataset, then fits other copies of the regression measure on the same dataset. The weights of the instances are adjusted according to the error of the current estimation. The learning rate is the weight reduction factor of each weak learner. In this work, we set the learning rate to 0.8 and the number of iterations to 60.

\paragraph{PLSR}

Partial Least Squares Regression(PLSR) \citep{wold2001pls}  combines the advantages of multiple linear regression and principal component analysis \citep{jolliffe2005principal}. The principal component pairs are derived by calculating the correlation coefficient matrix. Several spectral bands with relatively highest contribution values are selected to reduce  noise. In this work, we use grid search to determine the principal component and train the network for 50,000 iterations.

\section{RESULTS}

In this section, we evaluate our techniques in SSC and firmness estimation. 

\subsection{Data Set Processing}

\begin{figure}[h!]
	\begin{center}
		\includegraphics[width=15cm]{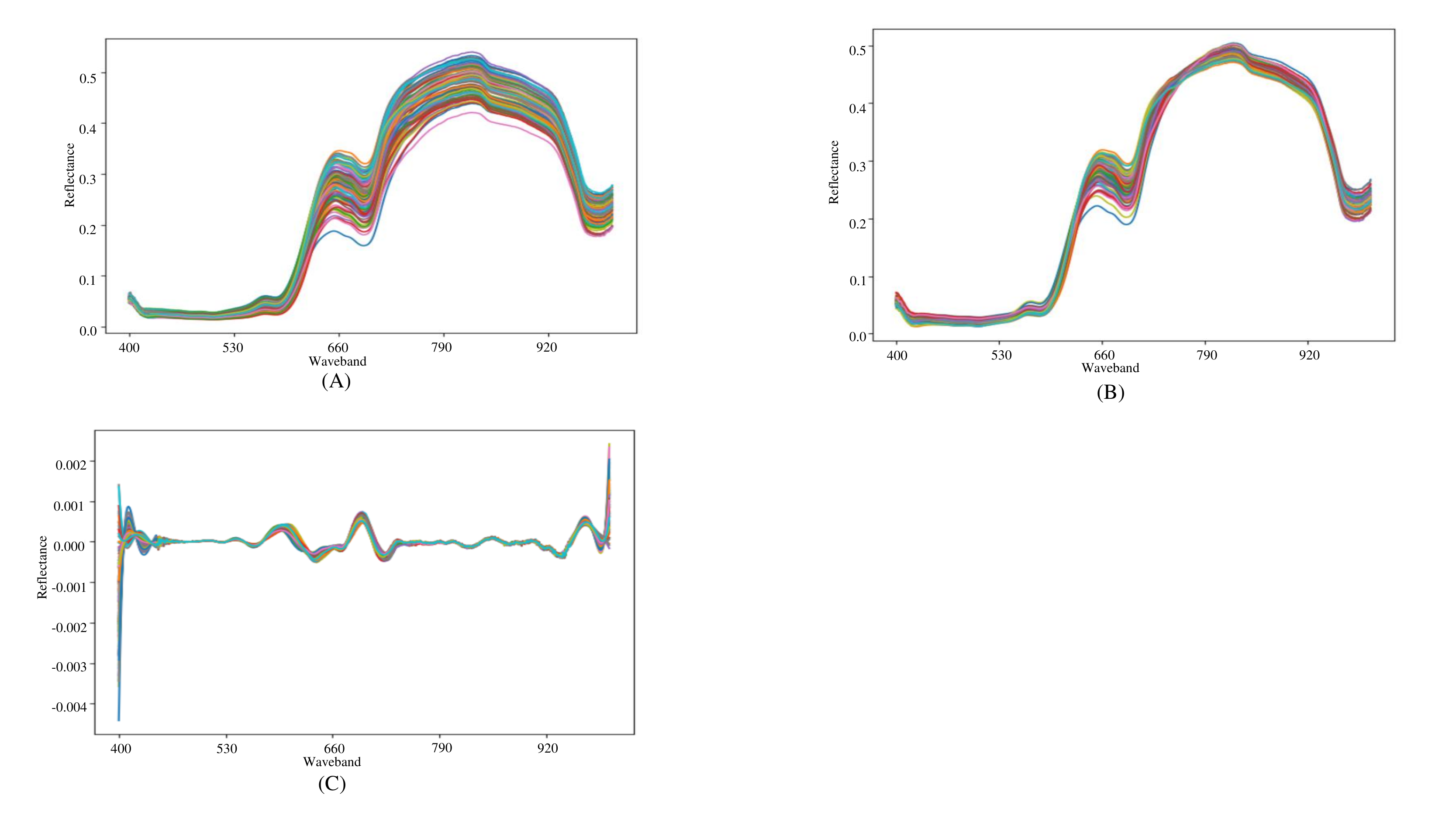}% This is a *.eps file
	\end{center}
	\caption{\textbf{(A)} Corrected spectral reflectance map, \textbf{(B)} MSC preprocessing, \textbf{(C)} Second-order differential preprocessing}\label{fig:5}
\end{figure}

The processed cherry tomato samples and the corresponding hyperspectral images are divided into training set, validation set, and test set with ratio of 7:1:2, respectively. We use varying data set size, with a small set if 50 samples and a large set of 200 samples. Fig.\ref{fig:5}A shows the reflectance spectra of 200 cherry tomato samples at 386-1004 nm. The spectral trends are similar for each sample since the reflection substances are the same. The cherry tomatoes have a strong absorption band at 400-550 nm due to the presence of carotenoids in ripe tomatoes\citep{ecarnot2013rapid}.The reflectance data are then processed using multiple scattering correction(MSC). As shown in Fig.\ref{fig:5}B, it can effectively reduce the noise and hence, smooth the curve. Finally, we use second order differentiation method ~\citep{ichige2006accurate} to process the smoothed reflectance data and discover clear peaks at locations of 580-590 nm, 680-690 nm, and 970-980 nm, as shown in Fig.\ref{fig:5}C. The three peaks are likely to be attributed to the combined effect of the second overtone of OH key, water, and tomato surface color\citep{li2013comparative,qiu2018variety}. Therefore, by proper processing, the variations in the spectral curves can reveal certain hidden information, such as SSC and water.

\subsection{Analysis}

Table \ref{tab:1} summarizes the distribution characteristics of SSC and firmness in different stages. The SSC and firmness measurements for the 50 and 200 samples are close to normally distributed around the mean values of 9.11$^\circ$ Brix, 9.04 N/cm2 and 8.72$^\circ$ Brix, 8.85 N/cm2 ,  standard deviations (SD) of 0.76, 1.35, and 0.66, 1.23, respectively.

\subsection{SSC Estimation Result}

Four machine learning models are implemented and compared with our proposed Con1dResNet network. We use ${R^2}$ and MSE as the evaluation metrics. They are calculated using the following equations.
$$
R^{2}=1-\frac{\sum_{i}\left(\widehat{y_{i}}-y_{i}\right)^{2}}{\sum_{i}\left(\overline{y_{i}}-y_{i}\right)^{2}}
$$
$$
M S E=\frac{1}{m} \sum_{i=1}^{m}\left(y_{i}-\widehat{y_{i}}\right)^{2}
$$
where ${\widehat{y_{i}}}$ is the estimated value, ${y_{i}}$ is the ground true value, and ${\overline{y_{i}}}$ is the mean  value. The optimal ${R^2}$ and MSE values are 1 and 0, respectively.

The algorithms are trained and run on a platform with an I7-8750H CPU and a 1060 GPU. They are programmed using python and tensorflow etc. The data sets are divided as described in Table \ref{tab:1}. The processed spectral data are used in the machine learning models while the raw spectral data are used in the Con1dResNet network. Since our deep learning model Con1dResNet can extract low to high dimensional features automatically, we use the original spectral data instead. We set Relu as the activation function, Adam as the optimizer, MSE as the loss function, the number of iterations to 50, and the batch size to 16. After 50 iterations of training, the loss decreases from 72.86 at the beginning to 0.01, indicating a convergence for the algorithm.

\begin{figure}[h!]
	\begin{center}
		\includegraphics[width=12.45cm]{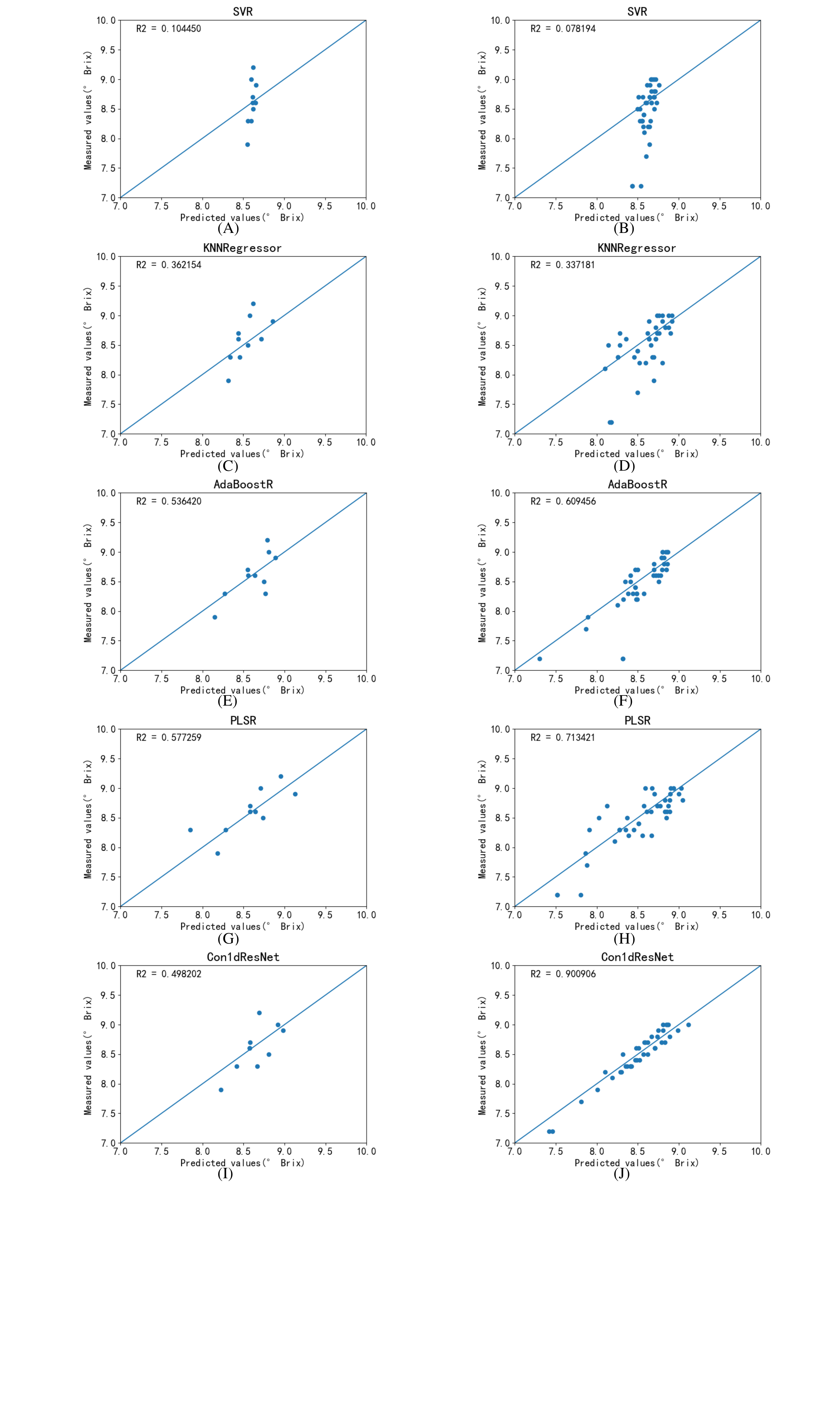}% This is a *.eps file
	\end{center}
	\caption{SSC estimation results for each model. \textbf{(A)} SVR estimation results on small sample data, \textbf{(B)} SVR estimation results on large sample data, \textbf{(C)} KNNR estimation results on small sample data, \textbf{(D)} KNNR estimation results on large sample data, \textbf{(E)} AdaBoostR estimation results on small sample data, \textbf{(F)} AdaBoostR estimation results on large sample data, \textbf{(G)} PLSR estimation results on small sample data, \textbf{(H)} PLSR estimation results on large sample data, \textbf{(I)} Con1dResNet estimation results on small sample data, \textbf{(J)} Con1dResNet estimation results on large sample data }\label{fig:6}
\end{figure}

The experimental results are shown in Fig.\ref{fig:6} and Table \ref{tab:2}.In general, the second-order differential processing outperforms MSC. However, since the SVR and KNNR models lack the ability of data dimensionality reduction, the noise caused by unwanted reflectance cannot be removed. When the data size increases, the amount of interference also rises. Thus, the ${R^2}$ value decreases as the data size increases. As expected, they have the worst performance with ${R^2}$ less than 0.4. For AdaBoostR, PLSR, and Con1dResNet models, ${R^2}$ values increase with increasing data sets size. For a relatively smaller data size, the PLSR model achieves the best performance, with ${R^2}$ of 0.577 and MSE of 0.055.  As the data size increases, the performance of the Con1dResNet model is improved significantly, with ${R^2}$ increasing from 0.498 to 0.901 (26.4\% better than the second best) and MSE decreasing from 0.065 to 0.018.

\subsection{Firmness Estimation Result}

\begin{figure}[h!]
	\begin{center}
		\includegraphics[width=15cm]{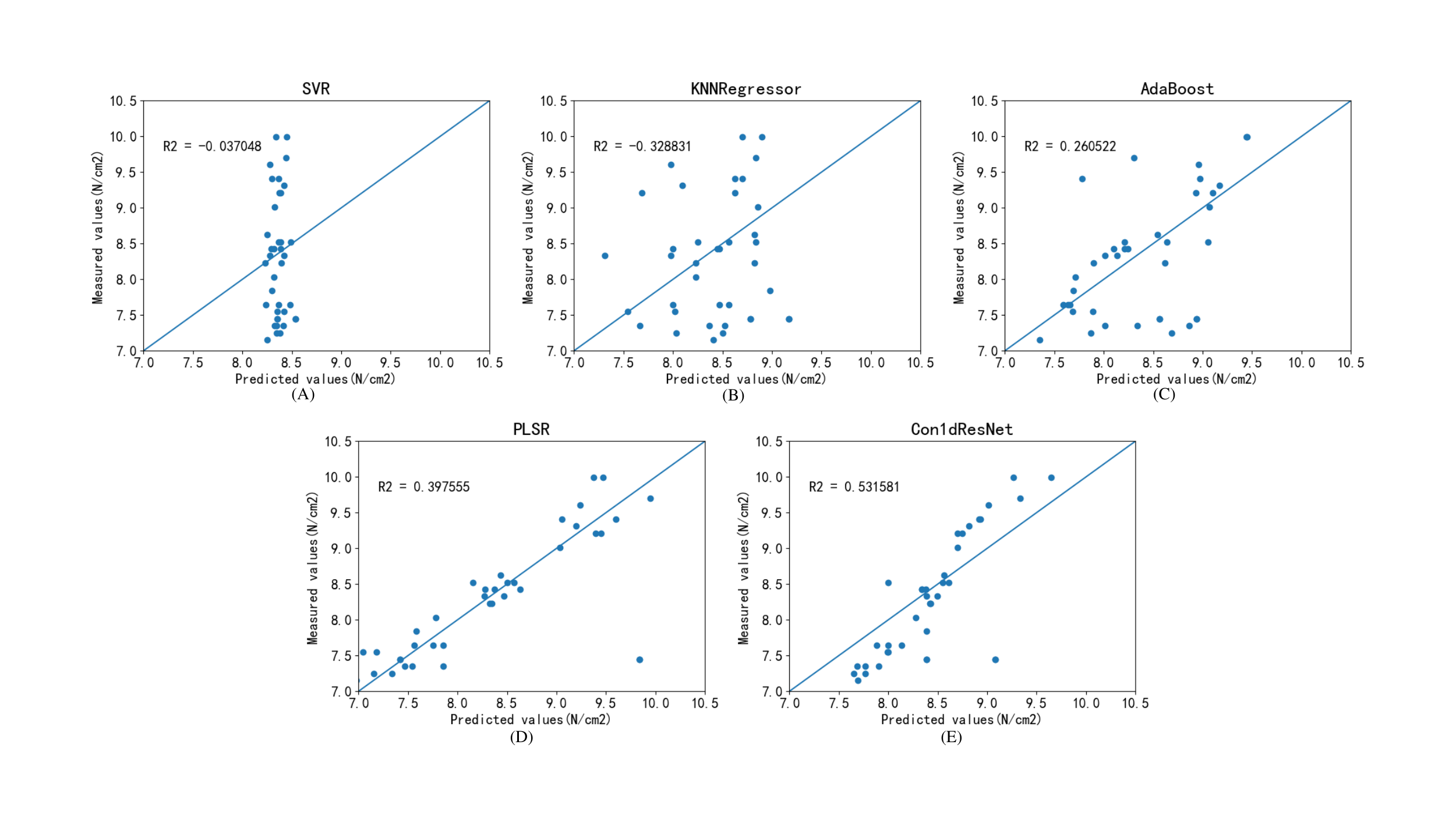}% This is a *.eps file
	\end{center}
	\caption{ Estimation results of firmness for each model on a large sample dataset}\label{fig:7}
\end{figure}

The same experimental setup is employed for firmness detection. As shown in Table.\ref{tab:3} and Fig.\ref{fig:7}, when MSC is employed for AdaBoost and PLSR, their ${R^2}$ values can be significantly improved \citep{wang2014effect}. Therefore, we choose MSC as the preprocessing method for AdaBoost and PLSR, and second-order difference as the preprocessing method for SVR and KNNR.   Although the method developed in this study has some advantages in data feature extraction compared with other methods, ${R^2}$ is still only 0.53, which does not achieve the accurate estimation standard. The ${R^2}$ of SVR and KNNR models is negative, which indicates the estimation accuracy is lower than the mean value.

\section{DISCUSSION}

The tomato flavor is important. SSC, which mainly consists of soluble sugars, can reflect the sweetness of cherry tomato. Hyperspectral imaging has been considered an effective technique for fruit SSC and firmness evaluation ~\citep{fan2015prediction,lu2004multispectral}. In this work, we discover a great estimation result for SSC estimation, while an inferior result for firmness.   

In general, the extracted spectral features ~\citep{guo2016hyperspectral} can derive excellent estimation results for large sample size. In that case, the deep learning based model performs significantly better than other machine learning based techniques since it can map the solution space to higher dimensions. For the PLSR \citep{wold2001pls} model,  it includes a principal component analysis component, which performs screening of band contribution values before training, and then selects 5-15 bands with relatively large contribution rates for regression. Thus, it can remove the influence of interfering bands \citep{he2019determination} and have more interference-free learning materials when the data size is large. Therefore, it can boost  ${R^2}$ when the data size increases. Combined with the removal of the invalid interference bands, the PLSR model becomes the state of art machine learning model. However, as the number of sample size increases, the  value of Con1dResNet model starts to gradually outperform the PLSR model due to the feature extraction ability of deep learning models. The experimental results demonstrate that Con1dResNet can significantly outperform the existing machine learning based techniques, with ${R^2}$ of 0.901 and MSE of 0.018. We believe that the experimental results of this work are also indicative for other horticultural crops.

For the hyperspectral images based tomato firmness, although it is reported that hyperspectral images can estimate fruit firmness\citep{fan2015prediction,lu2004multispectral}, our experimental results suggest otherwise. \cite{rahman2018hyperspectral} use PLSR to estimate tomato firmness using hyperspectral images in the 1000-1550 nm wavebands, and derive ${R^2}$ value of 0.6724. It is a little higher than our experiment due to the differences in the used hyperspectral wavebands  and the experimental environments. Therefore, in future work, for the estimation of firmness, we should explore a wider range of hyperspectral image wavebands, optimize the parameters for the firmness experiments, and improve the overall estimation accuracy.

\section{CONCLUSION}

In this work, we propose Con1dResNet, a deep learning based technique, to estimate the SSC and firmness of cherry tomatoes using hyperspectral images. With sufficient sample size, it can achieve better results than traditional machine learning methods. For SSC estimation, its ${R^2}$ value is 0.901, which is 26.4\% higher than PLSR, while its MSE is 0.018, which is 0.046 lower than PLSR. For Firmness estimation, its  ${R^2}$ value is 0.532, which is still 33.7\% better than PLSR. The results indicate that hyperspectral imaging combined with deep learning can significantly improve the cherry tomato SSC and firmness estimation accuracies.
\section*{Data Availability Statement}
The raw data supporting the conclusions of this article will be made available by the authors, without undue reservation.

\section*{Funding}
This work was supported by Provincial Key Research andDevelopment Program of Zhejiang (2021C02052).

\section*{Author Contributions}
YX, QC, YC, QX and ZS performed a conceptual, formal analysis of the study. QC, YX, ZC, and YC wrote the manuscript. YX, YC, QC, LZ, GZ, ZY, and QX designed the experiment. QC, LZ and ZS wrote the experimental code. YX, and QX verified the experimental results.  All authors contributed to the article and reviewed the manuscript.

%\section*{Acknowledgments}
%This is a short text to acknowledge the contributions of specific colleagues, institutions, or agencies that aided the efforts of the authors.

%\section*{Supplemental Data}
 %\href{http://home.frontiersin.org/about/author-guidelines#SupplementaryMaterial}{Supplementary Material} should be uploaded separately on submission, if there are Supplementary Figures, please include the caption in the same file as the figure. LaTeX Supplementary Material templates can be found in the Frontiers LaTeX folder.

\bibliographystyle{frontiersinSCNS_ENG_HUMS} 
%\bibliography{references}

\begin{thebibliography}{53}
\providecommand{\natexlab}[1]{#1}
\expandafter\ifx\csname urlstyle\endcsname\relax
  \providecommand{\doi}[1]{doi:\discretionary{}{}{}#1}\else
  \providecommand{\doi}{doi:\discretionary{}{}{}\begingroup
  \urlstyle{rm}\Url}\fi
\providecommand{\selectlanguage}[1]{\relax}
\providecommand{\bibAnnoteFile}[1]{%
  \IfFileExists{#1}{\begin{quotation}\noindent\textsc{Key:} #1\\
  \textsc{Annotation:}\ \input{#1}\end{quotation}}{}}
\providecommand{\bibAnnote}[2]{%
  \begin{quotation}\noindent\textsc{Key:} #1\\
  \textsc{Annotation:}\ #2\end{quotation}}

\bibitem[{Abdulridha et~al.(2019)Abdulridha, Batuman, and
  Ampatzidis}]{abdulridha2019uav}
Abdulridha, J., Batuman, O., and Ampatzidis, Y. (2019).
\newblock Uav-based remote sensing technique to detect citrus canker disease
  utilizing hyperspectral imaging and machine learning.
\newblock \emph{Remote Sensing} 11, 1373.
\newblock Doi:{ \href{https://doi.org/10.3390/rs11111373}{10.3390/rs11111373}}
\bibAnnoteFile{abdulridha2019uav}

\bibitem[{Ad{\~a}o et~al.(2017)Ad{\~a}o, Hru{\v{s}}ka, P{\'a}dua, Bessa, Peres,
  Morais et~al.}]{adao2017hyperspectral}
Ad{\~a}o, T., Hru{\v{s}}ka, J., P{\'a}dua, L., Bessa, J., Peres, E., Morais,
  R., et~al. (2017).
\newblock Hyperspectral imaging: A review on uav-based sensors, data processing
  and applications for agriculture and forestry.
\newblock \emph{Remote Sensing} 9, 1110.
\newblock Doi:{ \href{https://doi.org/10.3390/rs9111110}{10.3390/rs9111110}}
\bibAnnoteFile{adao2017hyperspectral}

\bibitem[{Afsah-Hejri et~al.(2019)Afsah-Hejri, Hajeb, Ara, and
  Ehsani}]{afsah2019comprehensive}
Afsah-Hejri, L., Hajeb, P., Ara, P., and Ehsani, R.~J. (2019).
\newblock A comprehensive review on food applications of terahertz spectroscopy
  and imaging.
\newblock \emph{Comprehensive reviews in food science and food safety} 18,
  1563--1621.
\newblock Doi:{
  \href{https://doi.org/10.1111/1541-4337.12490}{10.1111/1541-4337.12490}}
\bibAnnoteFile{afsah2019comprehensive}

\bibitem[{Beckles(2012)}]{beckles2012factors}
Beckles, D.~M. (2012).
\newblock Factors affecting the postharvest soluble solids and sugar content of
  tomato (solanum lycopersicum l.) fruit.
\newblock \emph{Postharvest Biology and Technology} 63, 129--140.
\newblock Doi:{
  \href{https://doi.org/10.1016/j.postharvbio.2011.05.016}{10.1016/j.postharvbio.2011.05.016}}
\bibAnnoteFile{beckles2012factors}

\bibitem[{Burger and Geladi(2005)}]{burger2005hyperspectral}
Burger, J. and Geladi, P. (2005).
\newblock Hyperspectral nir image regression part i: calibration and
  correction.
\newblock \emph{Journal of Chemometrics: A Journal of the Chemometrics Society}
  19, 355--363.
\newblock Doi:{ \href{https://doi.org/10.1002/cem.938}{10.1002/cem.938}}
\bibAnnoteFile{burger2005hyperspectral}

\bibitem[{Burges(1998)}]{burges1998tutorial}
Burges, C.~J. (1998).
\newblock A tutorial on support vector machines for pattern recognition.
\newblock \emph{Data mining and knowledge discovery} 2, 121--167.
\newblock Doi:{
  \href{https://doi.org/10.1023/A:1009715923555}{10.1023/A:1009715923555}}
\bibAnnoteFile{burges1998tutorial}

\bibitem[{Castro-Neto et~al.(2009)Castro-Neto, Jeong, Jeong, and
  Han}]{castro2009online}
Castro-Neto, M., Jeong, Y.-S., Jeong, M.-K., and Han, L.~D. (2009).
\newblock Online-svr for short-term traffic flow prediction under typical and
  atypical traffic conditions.
\newblock \emph{Expert systems with applications} 36, 6164--6173.
\newblock Doi:{
  \href{https://doi.org/10.1016/j.eswa.2008.07.069}{10.1016/j.eswa.2008.07.069}}
\bibAnnoteFile{castro2009online}

\bibitem[{Chen et~al.(2018)Chen, Chen, Liu, Wu, Wang, Ouyang
  et~al.}]{chen2018application}
Chen, Q., Chen, M., Liu, Y., Wu, J., Wang, X., Ouyang, Q., et~al. (2018).
\newblock Application of ft-nir spectroscopy for simultaneous estimation of
  taste quality and taste-related compounds content of black tea.
\newblock \emph{Journal of food science and technology} 55, 4363--4368.
\newblock Doi:{
  \href{https://doi.org/10.1007/s13197-018-3353-1}{10.1007/s13197-018-3353-1}}
\bibAnnoteFile{chen2018application}

\bibitem[{Du et~al.(2016)Du, Cai, Wang, and Zhang}]{du2016overview}
Du, X., Cai, Y., Wang, S., and Zhang, L. (2016).
\newblock Overview of deep learning.
\newblock In \emph{2016 31st Youth Academic Annual Conference of Chinese
  Association of Automation (YAC)} (IEEE), 159--164.
\newblock Doi:{
  \href{https://doi.org/10.1109/YAC.2016.7804882}{10.1109/YAC.2016.7804882}}
\bibAnnoteFile{du2016overview}

\bibitem[{Ecarnot et~al.(2013)Ecarnot, B{\k{a}}czyk, Tessarotto, and
  Chervin}]{ecarnot2013rapid}
Ecarnot, M., B{\k{a}}czyk, P., Tessarotto, L., and Chervin, C. (2013).
\newblock Rapid phenotyping of the tomato fruit model, micro-tom, with a
  portable vis--nir spectrometer.
\newblock \emph{Plant physiology and biochemistry} 70, 159--163.
\newblock Doi:{
  \href{https://doi.org/10.1016/j.plaphy.2013.05.019}{10.1016/j.plaphy.2013.05.0198}}
\bibAnnoteFile{ecarnot2013rapid}

\bibitem[{Elmasry et~al.(2012)Elmasry, Kamruzzaman, Sun, and
  Allen}]{elmasry2012principles}
Elmasry, G., Kamruzzaman, M., Sun, D.-W., and Allen, P. (2012).
\newblock Principles and applications of hyperspectral imaging in quality
  evaluation of agro-food products: a review.
\newblock \emph{Critical reviews in food science and nutrition} 52, 999--1023.
\newblock Doi:{
  \href{https://doi.org/10.1080/10408398.2010.543495}{10.1080/10408398.2010.543495}}
\bibAnnoteFile{elmasry2012principles}

\bibitem[{Esteva et~al.(2019)Esteva, Robicquet, Ramsundar, Kuleshov, DePristo,
  Chou et~al.}]{esteva2019guide}
Esteva, A., Robicquet, A., Ramsundar, B., Kuleshov, V., DePristo, M., Chou, K.,
  et~al. (2019).
\newblock A guide to deep learning in healthcare.
\newblock \emph{Nature medicine} 25, 24--29.
\newblock Doi:{
  \href{https://doi.org/10.1038/s41591-018-0316-z}{10.1038/s41591-018-0316-z}}
\bibAnnoteFile{esteva2019guide}

\bibitem[{Fan et~al.(2015)Fan, Huang, Guo, Zhang, and Zhao}]{fan2015prediction}
Fan, S., Huang, W., Guo, Z., Zhang, B., and Zhao, C. (2015).
\newblock Prediction of soluble solids content and firmness of pears using
  hyperspectral reflectance imaging.
\newblock \emph{Food analytical methods} 8, 1936--1946.
\newblock Doi:{
  \href{https://doi.org/10.1007/s12161-014-0079-1}{10.1007/s12161-014-0079-1}}
\bibAnnoteFile{fan2015prediction}

\bibitem[{{FAO}(2021)}]{Fao}
[Dataset] {FAO} (2021).
\newblock Tomato growth volume.
\newblock \url{https://www.fao.org/faostat/en/#data/QCL/visualize}.
\newblock Accessed December 27, 2021
\bibAnnoteFile{Fao}

\bibitem[{Freund et~al.(1999)Freund, Schapire, and Abe}]{freund1999short}
Freund, Y., Schapire, R., and Abe, N. (1999).
\newblock A short introduction to boosting.
\newblock \emph{Journal-Japanese Society For Artificial Intelligence} 14, 1612
\bibAnnoteFile{freund1999short}

\bibitem[{Fu et~al.(2021)Fu, Zhou, Scaboo, and Niu}]{fu2021nondestructive}
Fu, D., Zhou, J., Scaboo, A.~M., and Niu, X. (2021).
\newblock Nondestructive phenotyping fatty acid trait of single soybean seeds
  using reflective hyperspectral imagery.
\newblock \emph{Journal of Food Process Engineering} , e13759Doi:{
  \href{https://doi.org/10.1111/jfpe.13759}{10.1111/jfpe.13759}}
\bibAnnoteFile{fu2021nondestructive}

\bibitem[{Gao et~al.(2020)Gao, Liu, Yang, Zhang, Wang, Zhang
  et~al.}]{gao2020wrky}
Gao, Y.-F., Liu, J.-K., Yang, F.-M., Zhang, G.-Y., Wang, D., Zhang, L., et~al.
  (2020).
\newblock The wrky transcription factor wrky8 promotes resistance to pathogen
  infection and mediates drought and salt stress tolerance in solanum
  lycopersicum.
\newblock \emph{Physiologia plantarum} 168, 98--117.
\newblock Doi:{ \href{https://doi.org/10.1111/ppl.12978}{10.1111/ppl.12978}}
\bibAnnoteFile{gao2020wrky}

\bibitem[{Guo et~al.(2016{\natexlab{a}})Guo, Liu, Kong, He, Lou
  et~al.}]{guo2016hyperspectral}
Guo, C., Liu, F., Kong, W., He, Y., Lou, B., et~al. (2016{\natexlab{a}}).
\newblock Hyperspectral imaging analysis for ripeness evaluation of strawberry
  with support vector machine.
\newblock \emph{Journal of Food Engineering} 179, 11--18.
\newblock Doi:{
  \href{https://doi.org/10.1016/j.jfoodeng.2016.01.002}{10.1016/j.jfoodeng.2016.01.002}}
\bibAnnoteFile{guo2016hyperspectral}

\bibitem[{Guo et~al.(2016{\natexlab{b}})Guo, Liu, Oerlemans, Lao, Wu, and
  Lew}]{guo2016deep}
Guo, Y., Liu, Y., Oerlemans, A., Lao, S., Wu, S., and Lew, M.~S.
  (2016{\natexlab{b}}).
\newblock Deep learning for visual understanding: A review.
\newblock \emph{Neurocomputing} 187, 27--48.
\newblock Doi:{
  \href{https://doi.org/10.1016/j.neucom.2015.09.116}{10.1016/j.neucom.2015.09.116}}
\bibAnnoteFile{guo2016deep}

\bibitem[{He et~al.(2016)He, Zhang, Ren, and Sun}]{he2016identity}
He, K., Zhang, X., Ren, S., and Sun, J. (2016).
\newblock Identity mappings in deep residual networks.
\newblock In \emph{European conference on computer vision} (Springer), 630--645
\bibAnnoteFile{he2016identity}

\bibitem[{He et~al.(2019)He, Zhao, Zhang, Sun, and Li}]{he2019determination}
He, Y., Zhao, Y., Zhang, C., Sun, C., and Li, X. (2019).
\newblock Determination of {\ss}-carotene and lutein in green tea using fourier
  transform infrared spectroscopy.
\newblock \emph{Transactions of the ASABE} 62, 75--81.
\newblock Doi:{
  \href{https://doi.org/10.13031/trans.12839}{10.13031/trans.12839}}
\bibAnnoteFile{he2019determination}

\bibitem[{Hossain et~al.(2018)Hossain, Al-Hammadi, and
  Muhammad}]{hossain2018automatic}
Hossain, M.~S., Al-Hammadi, M., and Muhammad, G. (2018).
\newblock Automatic fruit classification using deep learning for industrial
  applications.
\newblock \emph{IEEE Transactions on Industrial Informatics} 15, 1027--1034.
\newblock Doi:{
  \href{https://doi.org/10.1109/TII.2018.2875149}{10.1109/TII.2018.2875149}}
\bibAnnoteFile{hossain2018automatic}

\bibitem[{Huang et~al.(2017)Huang, Lu, and Chen}]{huang2017development}
Huang, Y., Lu, R., and Chen, K. (2017).
\newblock Development of a multichannel hyperspectral imaging probe for
  property and quality assessment of horticultural products.
\newblock \emph{Postharvest Biology and Technology} 133, 88--97.
\newblock Doi:{
  \href{https://doi.org/10.1016/j.postharvbio.2017.07.009}{10.1016/j.postharvbio.2017.07.009}}
\bibAnnoteFile{huang2017development}

\bibitem[{Ichige et~al.(2006)Ichige, Ishikawa, and Arai}]{ichige2006accurate}
Ichige, K., Ishikawa, Y., and Arai, H. (2006).
\newblock Accurate direction-of-arrival estimation using second-order
  differential of music spectrum.
\newblock In \emph{2006 International Symposium on Intelligent Signal
  Processing and Communications} (IEEE), 995--998.
\newblock Doi:{
  \href{https://doi.org/10.1109/ISPACS.2006.364805}{10.1109/ISPACS.2006.364805}}
\bibAnnoteFile{ichige2006accurate}

\bibitem[{Jeyaraj and Nadar(2019)}]{jeyaraj2019computer}
Jeyaraj, P.~R. and Nadar, E. R.~S. (2019).
\newblock Computer-assisted medical image classification for early diagnosis of
  oral cancer employing deep learning algorithm.
\newblock \emph{Journal of cancer research and clinical oncology} 145,
  829--837.
\newblock Doi:{
  \href{https://doi.org/10.1007/s00432-018-02834-7}{10.1007/s00432-018-02834-7}}
\bibAnnoteFile{jeyaraj2019computer}

\bibitem[{Ji et~al.(2019)Ji, Sun, Li, and Ye}]{ji2019detection}
Ji, Y., Sun, L., Li, Y., and Ye, D. (2019).
\newblock Detection of bruised potatoes using hyperspectral imaging technique
  based on discrete wavelet transform.
\newblock \emph{Infrared Physics \& Technology} 103, 103054.
\newblock Doi:{
  \href{https://doi.org/10.1016/j.infrared.2019.103054}{10.1016/j.infrared.2019.103054}}
\bibAnnoteFile{ji2019detection}

\bibitem[{Jiang and Chen(2015)}]{jiang2015chemometric}
Jiang, H. and Chen, Q. (2015).
\newblock Chemometric models for the quantitative descriptive sensory
  properties of green tea (camellia sinensis l.) using fourier transform near
  infrared (ft-nir) spectroscopy.
\newblock \emph{Food Analytical Methods} 8, 954--962.
\newblock Doi:{
  \href{https://doi.org/10.1007/s12161-014-9978-4}{10.1007/s12161-014-9978-4}}
\bibAnnoteFile{jiang2015chemometric}

\bibitem[{Jolliffe(2005)}]{jolliffe2005principal}
Jolliffe, I. (2005).
\newblock Principal component analysis.
\newblock \emph{Encyclopedia of statistics in behavioral science} Doi:{
  \href{https://doi.org/10.1002/0470013192.bsa501}{10.1002/0470013192.bsa501}}
\bibAnnoteFile{jolliffe2005principal}

\bibitem[{Kamilaris and Prenafeta-Bold{\'u}(2018)}]{kamilaris2018deep}
Kamilaris, A. and Prenafeta-Bold{\'u}, F.~X. (2018).
\newblock Deep learning in agriculture: A survey.
\newblock \emph{Computers and electronics in agriculture} 147, 70--90.
\newblock Doi:{
  \href{https://doi.org/10.1016/j.compag.2018.02.016}{10.1016/j.compag.2018.02.016}}
\bibAnnoteFile{kamilaris2018deep}

\bibitem[{Kiranyaz et~al.(2015)Kiranyaz, Ince, Hamila, and
  Gabbouj}]{kiranyaz2015convolutional}
Kiranyaz, S., Ince, T., Hamila, R., and Gabbouj, M. (2015).
\newblock Convolutional neural networks for patient-specific ecg
  classification.
\newblock In \emph{2015 37th Annual International Conference of the IEEE
  Engineering in Medicine and Biology Society (EMBC)} (IEEE), 2608--2611.
\newblock Doi:{
  \href{https://doi.org/10.1109/EMBC.2015.7318926}{10.1109/EMBC.2015.7318926}}
\bibAnnoteFile{kiranyaz2015convolutional}

\bibitem[{Kong et~al.(2014)Kong, Liu, Zhang, Bao, Yu, and He}]{kong2014fast}
Kong, W., Liu, F., Zhang, C., Bao, Y., Yu, J., and He, Y. (2014).
\newblock Fast detection of peroxidase (pod) activity in tomato leaves which
  infected with botrytis cinerea using hyperspectral imaging.
\newblock \emph{Spectrochimica Acta Part A: Molecular and Biomolecular
  Spectroscopy} 118, 498--502.
\newblock Doi:{
  \href{https://doi.org/10.1016/j.saa.2013.09.009}{10.1016/j.saa.2013.09.009}}
\bibAnnoteFile{kong2014fast}

\bibitem[{Li et~al.(2013)Li, Huang, Zhao, and Zhang}]{li2013comparative}
Li, J., Huang, W., Zhao, C., and Zhang, B. (2013).
\newblock A comparative study for the quantitative determination of soluble
  solids content, ph and firmness of pears by vis/nir spectroscopy.
\newblock \emph{Journal of Food Engineering} 116, 324--332.
\newblock Doi:{
  \href{https://doi.org/10.1016/j.jfoodeng.2012.11.007}{10.1016/j.jfoodeng.2012.11.007}}
\bibAnnoteFile{li2013comparative}

\bibitem[{Li et~al.(2016)Li, Tian, Huang, Zhang, and Fan}]{li2016application}
Li, J., Tian, X., Huang, W., Zhang, B., and Fan, S. (2016).
\newblock Application of long-wave near infrared hyperspectral imaging for
  measurement of soluble solid content (ssc) in pear.
\newblock \emph{Food Analytical Methods} 9, 3087--3098.
\newblock Doi:{
  \href{https://doi.org/10.1007/s12161-016-0498-2}{10.1007/s12161-016-0498-2}}
\bibAnnoteFile{li2016application}

\bibitem[{Lowe et~al.(2017)Lowe, Harrison, and French}]{lowe2017hyperspectral}
Lowe, A., Harrison, N., and French, A.~P. (2017).
\newblock Hyperspectral image analysis techniques for the detection and
  classification of the early onset of plant disease and stress.
\newblock \emph{Plant methods} 13, 1--12.
\newblock Doi:{
  \href{https://doi.org/10.1186/s13007-017-0233-z}{10.1186/s13007-017-0233-z}}
\bibAnnoteFile{lowe2017hyperspectral}

\bibitem[{Lu(2004)}]{lu2004multispectral}
Lu, R. (2004).
\newblock Multispectral imaging for predicting firmness and soluble solids
  content of apple fruit.
\newblock \emph{Postharvest Biology and Technology} 31, 147--157.
\newblock Doi:{
  \href{https://doi.org/10.1016/j.postharvbio.2003.08.006}{10.1016/j.postharvbio.2003.08.006}}
\bibAnnoteFile{lu2004multispectral}

\bibitem[{Lu et~al.(2017)Lu, Huang, and Lu}]{lu2017innovative}
Lu, Y., Huang, Y., and Lu, R. (2017).
\newblock Innovative hyperspectral imaging-based techniques for quality
  evaluation of fruits and vegetables: A review.
\newblock \emph{Applied Sciences} 7, 189.
\newblock Doi:{ \href{https://doi.org/10.3390/app7020189}{10.3390/app7020189}}
\bibAnnoteFile{lu2017innovative}

\bibitem[{Medus et~al.(2021)Medus, Saban, Franc{\'e}s-V{\'\i}llora,
  Bataller-Mompe{\'a}n, and Rosado-Mu{\~n}oz}]{medus2021hyperspectral}
Medus, L.~D., Saban, M., Franc{\'e}s-V{\'\i}llora, J.~V., Bataller-Mompe{\'a}n,
  M., and Rosado-Mu{\~n}oz, A. (2021).
\newblock Hyperspectral image classification using cnn: Application to
  industrial food packaging.
\newblock \emph{Food Control} 125, 107962.
\newblock Doi:{
  \href{https://doi.org/10.1016/j.foodcont.2021.107962}{10.1016/j.foodcont.2021.107962}}
\bibAnnoteFile{medus2021hyperspectral}

\bibitem[{Paoletti et~al.(2019)Paoletti, Haut, Plaza, and
  Plaza}]{paoletti2019deep}
Paoletti, M., Haut, J., Plaza, J., and Plaza, A. (2019).
\newblock Deep learning classifiers for hyperspectral imaging: A review.
\newblock \emph{ISPRS Journal of Photogrammetry and Remote Sensing} 158,
  279--317.
\newblock Doi:{
  \href{https://doi.org/10.1016/j.isprsjprs.2019.09.006}{10.1016/j.isprsjprs.2019.09.006}}
\bibAnnoteFile{paoletti2019deep}

\bibitem[{Peng and Lu(2008)}]{peng2008analysis}
Peng, Y. and Lu, R. (2008).
\newblock Analysis of spatially resolved hyperspectral scattering images for
  assessing apple fruit firmness and soluble solids content.
\newblock \emph{Postharvest Biology and Technology} 48, 52--62.
\newblock Doi:{
  \href{https://doi.org/10.1016/j.postharvbio.2007.09.019}{10.1016/j.postharvbio.2007.09.019}}
\bibAnnoteFile{peng2008analysis}

\bibitem[{Qiu et~al.(2018)Qiu, Chen, Zhao, Zhu, He, and Zhang}]{qiu2018variety}
Qiu, Z., Chen, J., Zhao, Y., Zhu, S., He, Y., and Zhang, C. (2018).
\newblock Variety identification of single rice seed using hyperspectral
  imaging combined with convolutional neural network.
\newblock \emph{Applied Sciences} 8, 212.
\newblock Doi:{ \href{https://doi.org/10.3390/app8020212}{10.3390/app8020212}}
\bibAnnoteFile{qiu2018variety}

\bibitem[{Rahman et~al.(2017)Rahman, Kandpal, Lohumi, Kim, Lee, Mo
  et~al.}]{rahman2017nondestructive}
Rahman, A., Kandpal, L.~M., Lohumi, S., Kim, M.~S., Lee, H., Mo, C., et~al.
  (2017).
\newblock Nondestructive estimation of moisture content, ph and soluble solid
  contents in intact tomatoes using hyperspectral imaging.
\newblock \emph{Applied Sciences} 7, 109.
\newblock Doi:{ \href{https://doi.org/10.3390/app7010109}{10.3390/app7010109}}
\bibAnnoteFile{rahman2017nondestructive}

\bibitem[{Rahman et~al.(2018)Rahman, Park, Bae, and
  Cho}]{rahman2018hyperspectral}
Rahman, A., Park, E., Bae, H., and Cho, B.-K. (2018).
\newblock Hyperspectral imaging technique to evaluate the firmness and the
  sweetness index of tomatoes.
\newblock \emph{Korean Journal of Agricultural Science} 45, 823--837.
\newblock Doi:{
  \href{https://doi.org/10.7744/kjoas.20180075}{10.7744/kjoas.20180075}}
\bibAnnoteFile{rahman2018hyperspectral}

\bibitem[{Sainju et~al.(2003)Sainju, Dris, and Singh}]{sainju2003mineral}
Sainju, U.~M., Dris, R., and Singh, B. (2003).
\newblock Mineral nutrition of tomato.
\newblock \emph{Food, Agriculture \& Environment} 1, 176--183
\bibAnnoteFile{sainju2003mineral}

\bibitem[{Su et~al.(2021)Su, Zhang, Yan, Zhu, Zeng, Lu
  et~al.}]{su2021application}
Su, Z., Zhang, C., Yan, T., Zhu, J., Zeng, Y., Lu, X., et~al. (2021).
\newblock Application of hyperspectral imaging for maturity and soluble solids
  content determination of strawberry with deep learning approaches.
\newblock \emph{Frontiers in plant science} , 1897Doi:{
  \href{https://doi.org/10.3389/fpls.2021.736334}{10.3389/fpls.2021.736334}}
\bibAnnoteFile{su2021application}

\bibitem[{Wang et~al.(2014)Wang, Ji, and Gao}]{wang2014effect}
Wang, D.-M., Ji, J.-M., and Gao, H.-Z. (2014).
\newblock The effect of msc spectral pretreatment regions on near infrared
  spectroscopy calibration results.
\newblock \emph{Guang pu xue yu Guang pu fen xi= Guang pu} 34, 2387--2390
\bibAnnoteFile{wang2014effect}

\bibitem[{Weng et~al.(2019)Weng, Zhu, Zhang, Yuan, Zheng, Zhao
  et~al.}]{weng2019recent}
Weng, S., Zhu, W., Zhang, X., Yuan, H., Zheng, L., Zhao, J., et~al. (2019).
\newblock Recent advances in raman technology with applications in agriculture,
  food and biosystems: A review.
\newblock \emph{Artificial Intelligence in Agriculture} 3, 1--10.
\newblock Doi:{
  \href{https://doi.org/10.1016/j.aiia.2019.11.001}{10.1016/j.aiia.2019.11.001}}
\bibAnnoteFile{weng2019recent}

\bibitem[{Wold et~al.(2001)Wold, Sj{\"o}str{\"o}m, and Eriksson}]{wold2001pls}
Wold, S., Sj{\"o}str{\"o}m, M., and Eriksson, L. (2001).
\newblock Pls-regression: a basic tool of chemometrics.
\newblock \emph{Chemometrics and intelligent laboratory systems} 58, 109--130.
\newblock Doi:{
  \href{https://doi.org/10.1016/S0169-7439(01)00155-1}{10.1016/S0169-7439(01)00155-1}}
\bibAnnoteFile{wold2001pls}

\bibitem[{Xue(2010)}]{xue2010application}
Xue, L. (2010).
\newblock Application of idl and envi redevelopment in hyperspectral image
  preprocessing.
\newblock In \emph{International Conference on Computer and Computing
  Technologies in Agriculture} (Springer), 403--409
\bibAnnoteFile{xue2010application}

\bibitem[{Yao and Ruzzo(2006)}]{yao2006regression}
Yao, Z. and Ruzzo, W.~L. (2006).
\newblock A regression-based k nearest neighbor algorithm for gene function
  prediction from heterogeneous data.
\newblock In \emph{BMC bioinformatics} (BioMed Central), vol.~7, 1--11.
\newblock Doi:{
  \href{https://doi.org/10.1186/1471-2105-7-S1-S11}{10.1186/1471-2105-7-S1-S11}}
\bibAnnoteFile{yao2006regression}

\bibitem[{Yin et~al.(2019)Yin, Zhou, Chen, Han, Zheng, Younis
  et~al.}]{yin2019review}
Yin, L., Zhou, J., Chen, D., Han, T., Zheng, B., Younis, A., et~al. (2019).
\newblock A review of the application of near-infrared spectroscopy to rare
  traditional chinese medicine.
\newblock \emph{Spectrochimica Acta Part A: Molecular and Biomolecular
  Spectroscopy} 221, 117208.
\newblock Doi:{
  \href{https://doi.org/10.1016/j.saa.2019.117208}{10.1016/j.saa.2019.117208}}
\bibAnnoteFile{yin2019review}

\bibitem[{Yu and Deng(2010)}]{yu2010deep}
Yu, D. and Deng, L. (2010).
\newblock Deep learning and its applications to signal and information
  processing [exploratory dsp].
\newblock \emph{IEEE Signal Processing Magazine} 28, 145--154.
\newblock Doi:{
  \href{https://doi.org/10.1109/MSP.2010.939038}{10.1109/MSP.2010.939038}}
\bibAnnoteFile{yu2010deep}

\bibitem[{Zhou et~al.(2020)Zhou, Huang, Fan, Zhao, Liang, and
  Tian}]{zhou2020non}
Zhou, Q., Huang, W., Fan, S., Zhao, F., Liang, D., and Tian, X. (2020).
\newblock Non-destructive discrimination of the variety of sweet maize seeds
  based on hyperspectral image coupled with wavelength selection algorithm.
\newblock \emph{Infrared Physics \& Technology} 109, 103418.
\newblock Doi:{
  \href{https://doi.org/10.1016/j.infrared.2020.103418}{10.1016/j.infrared.2020.103418}}
\bibAnnoteFile{zhou2020non}

\bibitem[{Zou et~al.(2019)Zou, Shi, Guo, and Ye}]{zou2019object}
Zou, Z., Shi, Z., Guo, Y., and Ye, J. (2019).
\newblock Object detection in 20 years: A survey.
\newblock \emph{arXiv preprint arXiv:1905.05055}
\bibAnnoteFile{zou2019object}

\end{thebibliography}

%%% Make sure to upload the bib file along with the tex file and PDF
%%% Please see the test.bib file for some examples of references

\section*{Figure captions}

%%% Please be aware that for original research articles we only permit a combined number of 15 figures and tables, one figure with multiple subfigures will count as only one figure.
%%% Use this if adding the figures directly in the mansucript, if so, please remember to also upload the files when submitting your article
%%% There is no need for adding the file termination, as long as you indicate where the file is saved. In the examples below the files (logo1.eps and logos.eps) are in the Frontiers LaTeX folder
%%% If using *.tif files convert them to .jpg or .png
%%%  NB logo1.eps is required in the path in order to correctly compile front page header %%%

\begin{table}[h!]
	\caption{ Cherry tomato SSC and firmness dataset partitioning}
	\renewcommand\arraystretch{1.5}
	\begin{tabular}{cccccccccc}
		\toprule[2pt]
		\multirow{2}{*}{\begin{tabular}[c]{@{}c@{}}Sample\\ Size\end{tabular}}       & \multirow{2}{*}{Dataset} & \multicolumn{4}{c|}{SSC($\circ$ Brix)} & \multicolumn{4}{c}{firmness(N/cm2)} \\ \cline{3-10}
		&                          & MAX  & MIN & MEAN     & \multicolumn{1}{c|}{STD}      & MAX    & MIN   & MEAN    & STD      \\ \cline{2-10}
		\multirow{4}{*}{\begin{tabular}[c]{@{}c@{}}Small\end{tabular}} & Total(50)                & 10.800 & 8.000   & 9.114    & \multicolumn{1}{c|}{0.726} & 12.642 & 5.978 & 9.038 & 1.351 \\
		& Train Set(35)            & 10.800 & 8.000   & 9.129 & \multicolumn{1}{c|}{0.760} & 12.642 & 5.978 & 8.747  & 1.324 \\
		& Val Set(5)        & 10.400 & 8.700 & 9.320     & \multicolumn{1}{c|}{0.779} & 9.800    & 8.624 & 9.153  & 0.488 \\
		& Test Set(10)             & 9.200  & 7.900 & 8.600      & \multicolumn{1}{c|}{0.380} & 12.054 & 8.134 & 9.996   & 1.359 \\ \cline{2-10}
		\multirow{4}{*}{\begin{tabular}[c]{@{}c@{}}Large\end{tabular}} & Total(200)               & 11.100 & 7.200 & 8.719   & \multicolumn{1}{c|}{0.662} & 12.936 & 5.978 & 8.853 & 1.229 \\
		& Train Set(140)           & 11.100 & 7.200 & 8.790     & \multicolumn{1}{c|}{0.726} & 12.936 & 5.978 & 9.140  & 1.266 \\
		& Val Set(20)       & 9.200  & 7.800 & 8.500      & \multicolumn{1}{c|}{0.407} & 9.996  & 7.305  & 8.345  & 0.708 \\
		& Test Set(40)             & 9.000    & 7.200 & 8.455    & \multicolumn{1}{c|}{0.478} & 10.192 & 7.056 & 8.102 & 0.858 \\
		\bottomrule[2pt]
	\end{tabular}
	\label{tab:1}
\end{table}

\begin{table}[h!]
 \caption{${R^2}$ and MSE of estimated SSC for each model}
 \renewcommand\arraystretch{1.5}
 \begin{tabular}{lllll|ll}
  \toprule[2pt]
  \multirow{2}{*}{Sample   Size} & \multirow{2}{*}{Model} & \multirow{2}{*}{preprocessed} & \multicolumn{2}{l|}{Second-order   differential} & \multicolumn{2}{l}{MSC} \\ \cline{4-7}
  &                        &                               & ${R^2}$                       & MSE                    & ${R^2}$          & MSE        \\ \hline
  \multirow{5}{*}{Small(50)}     & SVR                    & \checkmark                             & 0.104                   & 0.116                  & 0.089      & 0.123      \\
  & KNNR                   & \checkmark                             & 0.362                   & 0.083                  & 0.289      & 0.096      \\
  & AdaBoost               & \checkmark                             & 0.536                   & 0.060                  & 0.502      & 0.068      \\
  & PLSR                   & \checkmark                            & 0.557                   & 0.055                  & 0.528      & 0.062      \\
  & Con1dResNet            & Original data                 & 0.498                   & 0.065                  & -           & -           \\ \hline
  \multirow{5}{*}{Large(200)}    & SVR                    & \checkmark                             & 0.078                   & 0.205                  & 0.075      & 0.207      \\
  & KNNR                   & \checkmark                             & 0.337                   & 0.147                  & 0.316      & 0.152      \\
  & AdaBoost               & \checkmark                             & 0.609                   & 0.089                  & 0.581      & 0.096      \\
  & PLSR                   & \checkmark                            & 0.713                   & 0.064                  & 0.710      & 0.067      \\
  & Con1dResNet            & Original data                 & 0.901                   & 0.018                  &  -          &  -   \\
  \bottomrule[2pt]
  \end{tabular}
 \label{tab:2}
\end{table}

\begin{table}[]
 \caption{${R^2}$ and MSE of estimated SSC for each model with all sample}
 \renewcommand\arraystretch{1.5}
 \begin{tabular}{llll|ll}
  \toprule[2pt]
  \multirow{2}{*}{Model} & \multirow{2}{*}{preprocessed} & \multicolumn{2}{l|}{Second-order   differential} & \multicolumn{2}{l}{MSC} \\ \cline{3-6}
  &                               & ${R^2}$                       & MSE                    & ${R^2}$           & MSE       \\ \hline
  SVR                    & \checkmark                             & -0.037                  & 1.108                  & -0.054      & 1.116     \\
  KNNR                   & \checkmark                             & -0.329                  & 1.251                  & -0.456      & 1.318     \\
  AdaBoost               & \checkmark                            & 0.217                   & 0.694                  & 0.261       & 0.675     \\
  PLSR                   & \checkmark                             & 0.384                   & 0.552                  & 0.398       & 0.548     \\
  Con1dResNet            & Original data                 & 0.532                   & 0.416                  &    -          &    -   \\
  \bottomrule[2pt]
 \end{tabular}
 \label{tab:3}
\end{table}
%\begin{figure}[h!]
%\begin{center}
%\includegraphics[width=15cm]{logos}
%\end{center}
%\caption{This is a figure with sub figures, \textbf{(A)} is one logo, \textbf{(B)} is a different logo.}\label{fig:8}
%\end{figure}

\end{document}